\pdfoutput=1

\documentclass[11pt]{article}

\usepackage{ACL2023}

\usepackage{times}
\usepackage{latexsym}

\usepackage[T1]{fontenc}

\usepackage[utf8]{inputenc}

\usepackage{microtype}

\usepackage{inconsolata}

\usepackage{color}
\usepackage{booktabs}
\usepackage{amsmath}
\usepackage{tabularx}
\usepackage{graphicx}
\usepackage{adjustbox}
\usepackage{multirow}
\usepackage{comment}
\usepackage{bbding}
\usepackage{pifont}
\usepackage{wasysym}
\usepackage{amssymb}
\usepackage{graphicx}

\title{Schema Graph-Guided Prompt for Multi-Domain Dialogue State Tracking}

\author{Ruolin Su ~\ Ting-Wei Wu \and Biing-Hwang Juang \\
         Georgia Institute of Technology \\ 
  \texttt{ruolinsu@gatech.edu, waynewu@gatech.edu, juang@ece.gatech.edu} \\}

\begin{document}
\maketitle
\begin{abstract}
Tracking dialogue states is an essential topic in task-oriented dialogue systems, which involve filling in the necessary information in pre-defined slots corresponding to a schema.
While general pre-trained language models have been shown effective in slot-filling, their performance is limited when applied to specific domains.
We propose a graph-based framework that learns domain-specific prompts by incorporating the dialogue schema.
Specifically, we embed domain-specific schema encoded by a graph neural network into the pre-trained language model, which allows for relations in the schema to guide the model for better adaptation to the specific domain.
Our experiments demonstrate that the proposed graph-based method outperforms other multi-domain DST approaches while using similar or fewer trainable parameters. 
We also conduct a comprehensive study of schema graph architectures, parameter usage, and module ablation that demonstrate the effectiveness of our model on multi-domain dialogue state tracking.
\end{abstract}

\section{Introduction}
Task-oriented dialogue systems have attracted great research interest due to their great potential for virtual assistants and other automated services.
As a crucial component in task-oriented dialogue systems, dialogue state tracking (DST) determines the user goals based on the previous dialogue turns, i.e. dialogue history. 
User goals are the tasks and purposes the user wants to accomplish through the dialogue and are typically represented as a set of pre-defined slot-value pairs corresponding to a domain-specific schema that consists of the required information to query the dialogue system. 
For instance, a typical \textit{``Find Flight''} service has a set of airline names and their relevant characteristics (e.g., \textit{Price, DepartureTime, DepartureDate}) as the dialogue constraints. The representations of user goals are a set of \textit{(slot, value)} pairs, such as \textit{(DepartureDate, next Wednesday)} and \textit{(DestinationCity, LAX)} for the \textit{``Find Flight''} service. 

\begin{figure}[t]
  \centering
  \includegraphics[width=\linewidth]{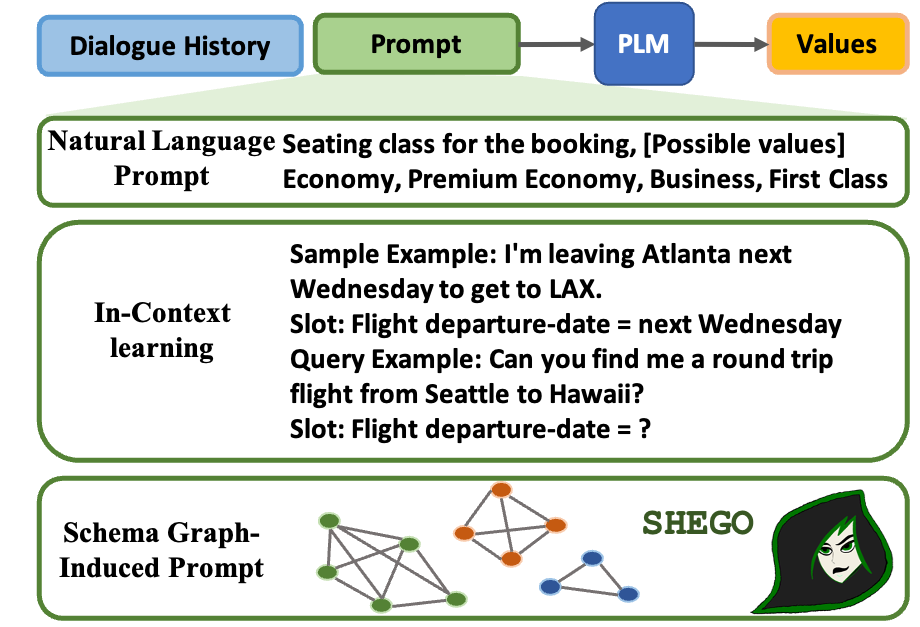}
  \caption{Illustration of prompt-based adaptation methods: natural language prompt, in-context learning, and schema graph-induced prompt.}
  \label{fig:SHEGO}
\end{figure}

Pre-trained language models (PLMs) achieved remarkable performance in task-oriented DST by fine-tuning with extensive task-oriented dialogues~\citep{zhang2019find, wu2019transferable, heck2020trippy, feng2020sequence}. 
However, their performance is at the cost of a large number of computational resources as the sizes of PLMs increase rapidly.
Recently, prompt-based adaption methods have been proposed, which freeze the PLM while only allowing a small number of parameters updated for downstream tasks~\cite{houlsby2019parameter, li2021prefix, lester2021power}. 
Such prompt-based adaption eases the computational cost of fine-tuning large language models per downstream task and thus improves the scalability of domain adaption.
Generally, there are two paradigms of prompt-based adaption methods for task-oriented DST: 
1) \textit{Natural Language Prompt} uses schema descriptions and possible values in natural language to draw domain-specific knowledge from the PLM~\citep{lee-etal-2021-dialogue, yang2022prompt,wang2022slot};
2) \textit{In-Context Learning} prepends related task examples to condition on the generated dialogue states~\cite{hu2022context, gupta2022show, venkateswaran2022district}.
However, the effectiveness of both paradigms highly depends on how well the conditional prompts fit in the input of PLM~\cite{lester2021power}.
Therefore, there is a growing demand for efficient multi-domain DST that can automatically update prompts and adeptly model domain-specific schema.

To address these issues, we propose a ScHEma Graph-guided prOmpt (SHEGO) paradigm based on graph neural networks (GNNs) ~\citep{scarselli2008graph, kipf2016semi} to incorporate the schema for domain-specific prompt learning,
All three paradigms are illustrated in Figure~\ref{fig:SHEGO}. 
Specifically, we model the multi-domain schema as a graph of slot tokens with their relations. 
GNNs are effective feature extractors using structural relations of domain-specific characteristics to condition on generating slot-value pairs and have been proved in previous work~\citep{joshi2021dialograph, feng2022dynamic}. 
Besides, we leverage several trainable tokens shared by all domains to adapt the pre-trained language model to DST.
Then, the dialogue history, slot tokens, and trainable shared tokens are combined in a row.
The schema graph represents domain- and service-specific information and relevant structural relations then are aggregated by ASAP pooling~\citep{ranjan2020asap}.
In other words, our graph-structured prompts are embedded through a GNN. 
In the end, we freeze the entire pre-trained model except for the GNN and embedding layers for dialogue history and the shared soft prompts.
Thus the limitation imposed on the parameters can help prevent overfitting to specific domains~\citep{lester2021power}.

Previous work has shown the efficiency and effectiveness of formulating DST in a sequence-to-sequence manner by predicting slot-value pairs all at once~\cite{wu2019transferable, lee-etal-2021-dialogue,yang2022prompt}. We thus formulate DST as masked span filling, which replaces desired values in the query with mask tokens and generates spans for masks as the output.
Such a value-masking formulation mimics the pattern in pre-training of language models which bridges the gap with downstream dialogue state tracking and has been proven effective in previous work~\citep{madotto2020continual,zhu-etal-2022-cpt4dst,wang2022slot}.
We use this to control the dialogue state generation with efficiency, as well as to leverage capabilities of natural language understanding in frozen PLM.


With T5-small~\citep{2020t5} as the backbone PLM, we evaluate SHEGO on two well-known DST benchmarks: Schema-Guided-Dialogue(SGD)~\citep{rastogi2020towards} and MultiWOZ 2.1~\citep{eric2019multiwoz}.
We show that our method achieves significant improvement in multi-domain dialogue state tracking on both benchmarks.
On the dataset level, SHEGO achieves state-of-the-art performance with a significant reduction of tunable parameters compared to existing models on SGD.
On the domain and service level, SHEGO has consistent gains by a margin on almost all domains, especially in domains with a larger number of slots. 

Our contribution is as follows: (1) We propose SHEGO, a graph-based prompt learning method that first incorporates slot relations in the schema and learns domain-aware prompts from domain-specific schema. (2) We conduct a comprehensive study of using GNN as an encoder for prompts for task-oriented DST and examine their differences and commonalities in terms of performance and characteristics. 
(3) Our experiments on SGD and MultiWOZ 2.1 demonstrate the effectiveness of the schema graph-guided prompts that enable domain adaptation with limited trainable parameters.

\begin{figure*}[htb]
  \centering
  \includegraphics[width=\textwidth]{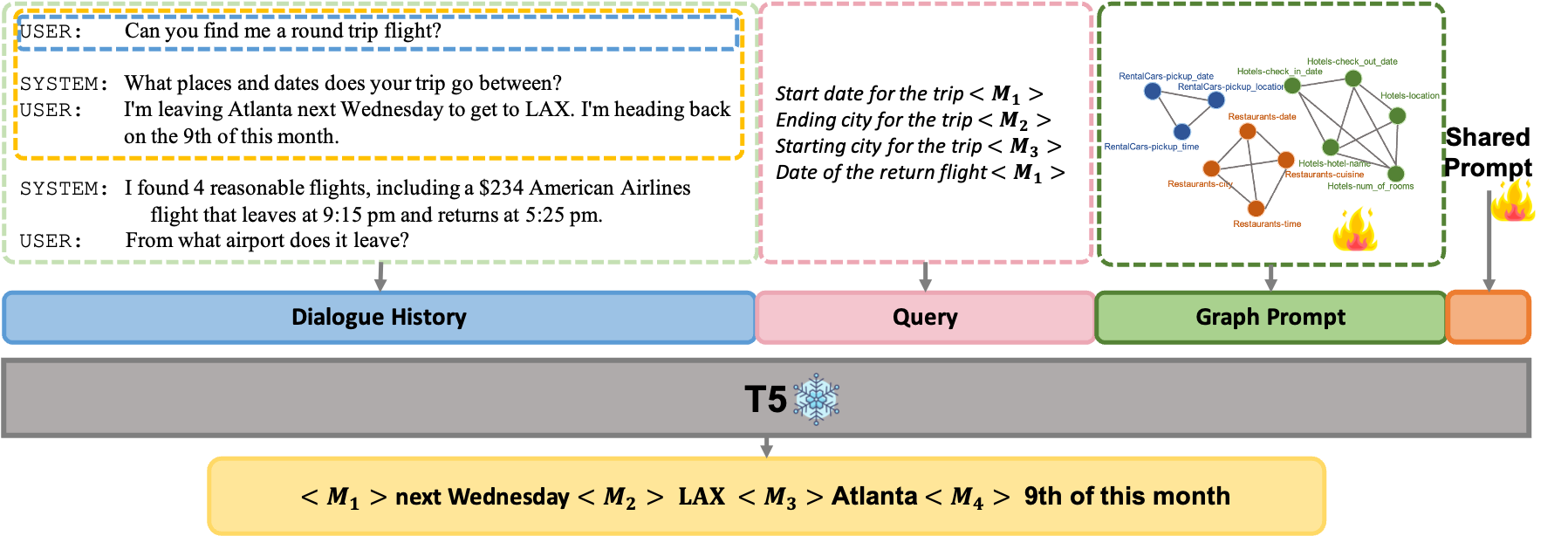}
  \caption{The overview of schema graph-induced prompt (SHEGO) architecture for multi-domain DST.} 
  \label{fig:model}
\end{figure*}

\section{Background}
\textbf{Prompt-based learning.}
Prompting is the approach of adding extra information for the model to condition when generating the output. 
As PLMs grow larger and larger, it becomes increasingly infeasible to perform conventional fine-tuning, where separate sets of the model parameters are modified for every single task. 
To reduce the effort in fine-tuning large PLMs and promote the scalability of domain adaptation,
there is a line of work that fixes the entire PLM and introduces a small number of new trainable parameters. Notable examples in this category include adapter~\citep{houlsby2019parameter, pfeiffer2020adapterfusion, karimi2021compacter}, prefix-tuning~\citep{li2021prefix} and prompt-tuning~\citep{lester2021power}, \textit{etc}.
Different from prompt design, which requires manual selection of prompt tokens from a fixed vocabulary, prompt tuning maintains a fixed set of special tokens, whereas only the embeddings of these prompt tokens are tuned.
Prompt tuning performs comparably or even better with conventional full-model tuning when downstream data are sufficient. Further studies improve prompt tuning by initializing it with an extensive training stage~\citep{gu2021ppt,vu2021spot}. 
In DST, prompts are used to help knowledge transfer in continual learning~\citep{madotto2020continual,zhu-etal-2022-cpt4dst} or address the capacity issue of domain adaptation~\citep{lee-etal-2021-dialogue}.

\textbf{Graph Neural Network for DST.}
Using structural relations between in-domain attributes is an effective way to construct and distill features.
GNN has been successfully applied in various dialogue applications. For example, \citep{ghosal-etal-2019-dialoguegcn} adopts a graph convolutional network (GCN) for utterance-level emotion recognition. \citet{Chen2018StructuredDP} modeled structured dialogue policy with GNN; \citet{qin2020dcr} and \citet{qin2020co} propose a joint framework leveraging graph attention network~\citep{velickovic2017graph} for both dialogue act recognition and sentiment classification.
Recently, DSGFNet~\citep{feng2022dynamic} leverages the schema graph encoder that fuses slot-domain and dialogue-aware relations for DST; DiCoS-DST~\citep{guo2022beyond} selects relevant turns dynamically by modeling structural slot dependency. 


\textbf{Dialogue State Tracking.} Research on DST evolves from conventional generative models to discriminative models, from fixed ontology to scalable scenarios for more complicated dialogue applications. 
Specifically, some previous works formulate state transitions during a dialog as a probabilistic process assuming fixed sets of dialogue states~\citep{young2013pomdp}. Recent work adopts slot-value pairs for dialogue state representations~\cite{Mrksic,rastogi2020towards,su2022act}, where slots can be drawn as pre-defined value sets or free-form span values.
For the former scenario, a line of works formulates DST as a classification problem assuming candidate values are fixed for each slot~\cite{wu2019transferable,liu2017end,AbhinavRastogiDilekHakkani-Tur2017} \vphantom{[GLAD, e2e belief tracker, multi-channel]}. 
However, these approaches are constrained by the pre-defined domain ontological sets that often fail in unknown values or scaling to new domains. \citet{Shi2017} and \citet{Zhong2018} \vphantom{[TRADE, scalable neural DST]} expand DST models to track slots whose value sets are not fixed. Recent achievements in neural machine reading shed light on tackling DST without extensive fine-tuning~\citep{lin-etal-2021-zero,su2023choice}.

\section{Method}
We introduce SHEGO, a prompt learning approach based on GNN—\textit{i.e.}, a graph convolutional neural network (GCN)—that learns graph prompt embeddings from the schema.
Figure 1 shows the architecture of our proposed method.
Specifically, the GCN layers encode relations among all slots in the dialogue schema into embeddings, which are then combined with embeddings of shared soft prompt tokens.
The PLM, \textit{i.e.} T5-small~\citep{2020t5}, and other parts of the input are fixed, whose weights are drawn from pre-training.
We elaborate on DST task formulation (Sec. \ref{sec:msf}), a GNN-based prompt encoder (Sec~\ref{sec:gpe}) and prompt training (Sec~\ref{sec:end2end}) on the following:

\subsection{Masked Slot Filling}
\label{sec:msf}
In dialogue state tracking, the schema defines each domain $\mathcal{D}_l$ and its corresponding slots $\mathcal{S}_l$ in a tuple $\{\mathcal{D}_l: d^{D_l}, \mathcal{S}_l\}$, where $\mathcal{S}_l = \{s_1:d_1, \dots, s_{n_l}:d_{n_l}\}$ and $d$ are descriptions for domains and slots.
The objective of DST is to output $Y = \{(s_1,v_1), \dots, (s_{n_l},v_{n_l})\}$ as slot-value pairs, given the dialogue history $H = \left[u_1^{sys},u_1^{usr},\dots,u_t^{sys},u_t^{usr}\right]$ as the concatenation of the system and user utterances in previous turns, 
where $t$ is the number of current turns in the dialogue.

We experiment with T5, which is an encoder-decoder model pre-trained to predict corrupted tokens in the input, using sentinel tokens as the hint~\citep{zhu-etal-2022-cpt4dst}.
To take after the objective in pre-training, we also substitute each value in $Y$ with a sentinel token. 
Specifically, we form a query for each input sequence by combining sentinel tokens of slots with their descriptions into a sequence:
\begin{equation}
Q_l=\left[q_1,\ldots q_{n_l}\right], \textit{where}\; q_j=d_j: \langle\mathcal{M}_j\rangle
\end{equation}
$d_j$ are descriptions of the slot, and $\langle \mathcal{M}_j \rangle$ are distinct sentinel tokens of the $j$-\textit{th} slot representing masks.
Such a combined sequence is then appended to dialogue history. 
DST is thus formulated as imputing $v$ in the output sequence $Y\sp{\prime}$ for each domain.
\begin{equation}
Y\sp{\prime} = \langle \mathcal{M}_1\rangle v_1\langle\mathcal{M}_2\rangle v_2 \dots \langle\mathcal{M}_{n_l}\rangle v_{n_l}
\end{equation}

To leverage the power of trainable prompt tokens and reduce the computational costs, we append $m=\sum_{l}\left|n_l\right|$ graph prompts $G$ and $p$ shared soft prompts $P$ to the input sequence.
\begin{equation}G={G}_1{G}_2\ldots{G}_m\end{equation}
\begin{equation}P={P}_1{P}_2\ldots{P}_p\end{equation}
In the following, we generate structural representations of prompts $G$ by the schema-graph encoder and train embeddings of soft prompt $P$.

\subsection{Schema Graph Prompt Encoder}
\label{sec:gpe}
Our schema graph prompt encoder is designed to model slot relations in the dialogue schema and extract their global characteristics.
 These structural representations are used to condense the dialogue representation and fit in the dialogue input to the frozen pre-trained language model.
 Below we describe the structure encoder for domain ontology.

Denoting distinct slots as $G^i$ of the $i$-\textit{th} dialogue and the corresponding adjacency matrix as $\mathcal{A}^i \in R^{m \times m}$, we model the slot level relations by building up an undirected graph, \textit{i.e.} $\{G^i, {\mathcal{A}}^{i}\}$,
where $m$ is the number of slots of all domains in the training data. 
The adjacency matrix is formulated as follows: Given two slots $u,v\in G^i$ and the connectivity of $\left(u,v\right)$ as $\mathcal{A}_{u,v}^{i}$, if $u, v$ belong to the same domain, we build an edge from a particular node of slots to another node. 
These edges incorporate slot-level semantics contained in the schema.

Moreover, we transform $G^i \rightarrow \Tilde{G}^i$ by masking $G^i$ with active (not none) slots in the $i$-\textit{th} dialogue. Specifically,
\begin{equation}
    \Tilde{G}_j^i = 
    \begin{cases}
    G_j^i& \textit{if $j$-\textit{th} slot active}\\
    0& \textit{otherwise}\\
    \end{cases}
\end{equation}
The motivation behind this is that active slots are contextual information for each dialogue turn in training. 
We hypothesize that this setup improves the diversity of input graph prompts, such that making adaptation to the task more promptly.

We use a hierarchical graph pooling-based encoder, which includes multiple layers of GCN each followed by adaptive structure-aware pooling (ASAP) layer~\citep{ranjan2020asap}. 
The GCN analyzes the input graph to create structurally informed representations of the nodes. ASAP then groups similar nodes and ranks the clusters, forming new nodes and edges from the top clusters. 
This process generates a structurally informed graph representation and allows us to utilize shared information from ontology in other domains.

In this way, the graph representation is learned and accumulated preserving the structural information. After each pooling step, the graph representation is summarized using the concatenation of the mean and max of the node representations. The summaries are then added and passed through fully connected layers to obtain the final structural representation of the domain ontology.

\subsection{End-to-End Training}
\label{sec:end2end}
We combine the dialogue history $H$, dialogue query $Q_l^i$, graph prompts $G$, and shared soft prompts $P$ together as the input to the frozen T5. 
Prompts $G$ and $P$ can be viewed as the flexible context that helps the fixed pre-trained model adapt to DST tasks in specific domains.
With the input augmented with trainable graph prompts and shared soft prompts, we train our model in an end-to-end manner.

Our model maximizes the conditional probability of $Y\sp{\prime}$ by training the parameters $\theta_P$ and $\theta_Q$ of the prompts, while all other parameters are fixed:
\begin{equation}
    \mathcal{L}_{\theta_P, \theta_Q} = - \sum_i^{\left|D_l\right|} \log p_{\theta_P, \theta_Q}(Y\sp{\prime}| H^i, Q_l^i, \Tilde{G}^i, P^i)
\end{equation}

\begin{table}[htb]
  \centering
  \begin{adjustbox}{width=\linewidth}
  \begin{tabular}{lcl}
    \toprule
    \textbf{Model} & Pretrained Model & \textit{Avg. JGA} \\
    \midrule
    SGD-baseline & BERT-base / 110M & 25.1 \\
    Seq2Seq-DU & BERT-base / 110M & 22.1 \\
    AdapterCL$^\dagger$ & GPT-2 / 1.5B & 39.7 \\
    Prompt-Tuning$^\ast$ & T5-small / 60M &  73.1 \\
    SHEGO & T5-small / 60M & \textbf{76.6}$_{\pm3.4}$ \\

    \bottomrule
  \end{tabular}
  \end{adjustbox}
   \caption{Accuracy on each domain on the SGD test set. AdapterCL$^\dagger$ is drawn from their best model's results, and Prompt-Tuning$^\ast$ is drawn from our re-implementation.}
 \label{tab:sgd-joint}
\end{table}

\begin{table*}
  \centering
  \begin{adjustbox}{width=\linewidth}
  \begin{tabular}{c|c|c|c|c|c|c|c|c|c|c}
    \toprule
    \textbf{Domain} & \multicolumn{2}{|c|}{\textbf{SGD-baseline}} & \multicolumn{2}{|c|}{\textbf{Seq2Seq-DU}} & \multicolumn{2}{|c}{\textbf{AdapterCL$^\dagger$}} & \multicolumn{2}{|c}{\textbf{Prompt-Tuning$^\ast$}} & \multicolumn{2}{|c}{\textbf{SHEGO}}\\
    \midrule
    \textit{Metric} & \textbf{JGA} & \textbf{AGA} & \textbf{JGA}& \textbf{AGA} & \textbf{JGA} & \textbf{AGA} & \textbf{JGA} & \textbf{AGA}& \textbf{JGA}& \textbf{AGA}\\
    \midrule
    \textit{Travel} & 41.5    &57.2     &44.9    &-     &39.6&58.9    &76.2&81.5          & 80.4$_{\pm9.2}$    &88.8$_{\pm4.5}$\\
    \textit{Weather} & 62.0    &76.4     &57.9    &-    &72.5&86.0       &95.5&97.1       & 97.0$_{\pm1.5}$    &98.1$_{\pm1.0}$\\
    \textit{RideSharing} & 17.0    &50.2      &67.0    &- &61.5&81.3    &91.0&97.4        &86.3$_{\pm3.7}$    &94.7$_{\pm1.3}$\\
    \textit{Homes} & 18.9    &72.7      &22.8    &-     &45.4&71.9     &81.9&91.0         & 82.6$_{\pm3.8}$    &94.9$_{\pm1.9}$\\
    \textit{Hotels} & 28.9    &58.2     &34.0    &-   &31.3&62.5   &73.6&91.5     & 80.4$_{\pm2.8}$    &93.3$_{\pm0.8}$\\
    \textit{Movies} & 37.8    &68.6     &43.9    &-   &26.9&40.4   &65.1&90.9     & 72.5$_{\pm3.6}$    &92.5$_{\pm2.0}$\\
    \textit{Services} & 40.9    &72.1   & 47.7    &-    &28.6&69.4 &74.3&89.6     & 77.2$_{\pm3.5}$   &91.4$_{\pm1.5}$\\
    \textit{Buses} &9.7    &50.9  &    16.8&-     &23.1&56.9 &78.9&94.4      &81.3$_{\pm3.2}$    & 94.9$_{\pm1.0}$\\
    \textit{Payment} & 11.5    &34.8   &7.2    &-   &58.8&87.7 &37.1&65.5      &$43.4_{\pm12.0}$    & $81.4_{\pm4.2}$\\
    \textit{Trains} & 13.6    &63.5     &16.8    &-   &24.4&62.3   &60.7&88.7      &65.8$_{\pm10.9}$    & 90.6$_{\pm2.7}$\\
    \textit{Music} & 15.5   &39.9     &12.3   &-    &21.4&61.0    &62.1&84.2       &64.7$_{\pm2.7}$     & 84.1$_{\pm3.0}$\\
    \textit{RentalCars} &8.6    &48.0   &6.25    &-   &20.5&61.7 &64.8&91.6      &72.6$_{\pm3.4}$    & 93.0$_{\pm1.3}$\\
    \textit{Restaurants} & 22.8    &55.8&13.0    &- &16.9&61.0     &67.0&92.3      &67.9$_{\pm3.2}$    & 92.3$_{\pm1.6}$\\
    \textit{Events} & 23.5    &57.9 &31.8  &-   &28.6&60.2   &77.6&94.2      &79.9$_{\pm2.6}$     & 95.1$_{\pm0.8}$\\
    \textit{Flights} & 23.9    &65.9&15.9    &-   &22.0&52.8  &77.6&95.5      &80.8$_{\pm4.0}$     & 96.3$_{\pm0.8}$\\
    \textit{Alarm} & 57.7    &1.8 &55.7    &-  &58.8&87.7   &87.8&90.5      &87.9$_{\pm1.6}$     & 93.2$_{\pm2.9}$\\

    \bottomrule
  \end{tabular}
 \end{adjustbox}
 \caption{Accuracy of our model and baselines on SGD dataset by domains. Our model outperforms baselines on joint and slot goal accuracies. AdapterCL$^\dagger$ is drawn from their best model's results, and Prompt-Tuning$^\ast$ is drawn from our re-implementation.}
 \label{tab:sgd-domain}

\end{table*}

\section{Experiment}
\subsection{Dataset}
We verify the effectiveness of our models on widely used benchmarks for multi-domain task-oriented dialogue state tracking: Schema-Guided-Dialogue (SGD)~\citep{rastogi2020towards} and MultiWOZ 2.1~\citep{eric2019multiwoz}.
The SGD dataset provides 38K training examples of 44 services over 19 domains, following pre-processing from~\citet{madotto2020continual}. 
The schema defines each slot and corresponding service, with a brief description of the slot and service.
We only consider dialogues of a single service and take \textit{Service}$_l$ as $D_l$ in our method following \citet{zhu-etal-2022-cpt4dst}.
We calculate the accuracy of several tasks from the same domain and present our presented results per domain to make a fair comparison.

We also evaluate models on MultiWOZ 2.1 and follow dataset setups in~\citet{wu2019transferable} that experiments with 5 most frequent domains in the dataset: \{\textit{restaurant, hotel, train, attraction, taxi}\} and 30 slots.
12K examples and slot-level descriptions are used for training.
Detailed statistics of datasets are presented in Table~\ref{tab:stat}.

\subsection{Metric}
We use joint goal accuracy (\textbf{JGA}) and average goal accuracy (\textbf{AGA}) to evaluate our models and baselines. 
Joint goal accuracy is the average accuracy of predicting all slot-values for a turn correctly, while average goal accuracy is the average accuracy of predicting the value of a slot correctly for active slots in the ground truth. 
A slot is called active if its value is NOT none in the ground truth dialogue state of the current turn. 
We compute JGA and AGA on SGD by domains and overall JGA on MultiWOZ datasets, reporting the mean and the standard deviation of 3 random runs. 
We also report \textit{Avg.}~JGA defined as the average of JGA across all $L$ domains: 
\begin{equation}\textit{Avg.}~\text{JGA} = \frac1{L} \sum_{l=1}^L \text{JGA}_{\mathcal{D}_l}\end{equation}

\subsection{Implementation detail}
We implement our approach based on T5-small (60M parameters)~\citep{raffel2020exploring} pre-trained language model whose hidden dimension is 512.  
Parameters inside of T5-small are fixed, with only trainable GNN layers and embeddings for shared prompts, at a batch size of 16 using AdamW ~\citep{loshchilov2018decoupled} optimizer. The initial learning rate is set to 0.01 with weight decay $5e-4$ for GNN layers and 0.5 without weight decay for T5. 
We use greedy decoding for the maximum length of 100 for all models.
Early stopping is adopted if validation performance does not improve for 5 consecutive epochs.
The number of prompts for the schema graph is equal to that of domain-slot pairs in all the schema. 
Tokens representing domain-slot pairs are augmented to the model's vocabulary and their embeddings are initialized with normal distribution.
The hidden dimension of the ASAP layer is 256 while its output layer has a size of 512.
We use embeddings that enumerate 100 additional tokens in the vocabulary as the additional tunable prompt initialization with the embedding size 512.

\subsection{Baseline}
We evaluate the performance of SHEGO for multi-domain DST compared with the following baseline models. 
\textbf{SGD-baseline}~\citep{rastogi2020towards} fine-tunes BERT to predict values for each slot. 
\textbf{Seq2Seq-DU}~\citep{feng2020sequence} employs BERT in the encoding of utterances and schema descriptions respectively and generates pointers in decoding.
\textbf{TRADE}~\citep{wu2019transferable} adopts slot gate and copy mechanism to track slot values mentioned in the dialogue history.
\textbf{TripPy}~\citep{heck2020trippy} enhances DST with a triple copy mechanism exploiting system inform memory as an extra input.
\textbf{TripPy-R}~\citep{heck2022robust} combines the copy mechanism in TripPy with a unified encoder and attention mechanism to improve slot matching.
\textbf{T5DST}~\citep{lee-etal-2021-dialogue} proposes schema-driven prompting to improve DST in low-source scenarios.
\textbf{MoNET}~\citep{zhang2022monet} leverages contrastive context matching to update and correct slot values.
\textbf{PPTOD}~\citep{su-etal-2022-multi} presents a unified plug-and-play model for DST with task names as prompts.
\textbf{SeKnow-PLM}~\citep{gao2022end} improves dialogue modeling grounded on semi-structured knowledge.
\textbf{DiSTRICT}~\citep{venkateswaran2022district} utilizes retrieved in-context examples to fine-tune the language model.
\textbf{DiCoS-DST}~\citep{guo2022beyond} explicitly models slot dependency to select relevant contents dynamically.
\textbf{AdapterCL}~\citep{madotto2020continual} learns an adapter to avoid catastrophic forgetting in continual learning.
\textbf{Prompt Tuning} for DST~\citep{zhu-etal-2022-cpt4dst} first proposes to use prompt tuning for DST focusing on using prompts to transfer knowledge in continual learning,

\section{Results and Analyses}
We show the effectiveness of our models on Schema-Guided-Dialogue(SGD) (Sec.~\ref{sec:sgd}) and MultiWOZ (Sec.~\ref{sec:mwoz}) compared with baselines, respectively.
And we show the model performance when the ASAP layers have different GNN architectures in Sec.~\ref{sec:architect}.
In Sec.~\ref{sec:abaltion}, we investigate our proposed approaches with a comprehensive ablation study.

\subsection{Results on SGD}
\label{sec:sgd}

\begin{table}
  \centering
  \begin{adjustbox}{width=\linewidth}
  \begin{tabular}{lcll}
    \toprule
    \textbf{Model} & Pretrained Model / \textit{\#Param.} & \textit{JGA} & \textit{AGA}\\
    \midrule
    SGD-baseline & BERT-base / 110M & 43.4 & - \\
    TRADE&  -                   &48.6 & 96.9\\
    Seq2Seq-DU & BERT-base / 110M & 56,1 & 91.1\\
    TripPy & BERT-base / 110M & 55.3 & -\\
    T5DST & T5-base  / 220M & 56.7 \\
    MoNET & BERT-base / 110M & 57.7 & 97.7 \\
    PPTOD & T5-large/770M & 57.5 & -\\
    SeKnow-PLM & GPT-2 / 1.5B & 58.5 & -\\
    Trippy-R & RoBERTa-base / 120M & 56.0 & -\\
    DiSTRICT & T5-small / 60M & 56.1 & -\\
    DiCoS-DST & ALBERT-large / 18M & 61.0 & 98.1\\
    \midrule \midrule
    SHEGO & T5-small / 60M & 59.0$_{\pm0.6}$ & 95.6$_{\pm0.1}$ \\

    \bottomrule
  \end{tabular}
     \end{adjustbox}
 \caption{Joint and slot goal accuracy on each domain on MultiWOZ2.1.}
 \label{tab:mwoz-joint}
\end{table}

Table~\ref{tab:sgd-joint} shows the evaluation results of using T5-small on the SGD test set. The results are averaged over three random seeds. Our proposed approaches with frozen T5-small achieve the state-of-the-art JGA by 3.5\%, even outperforming GPT-2 with 1.2B parameters. 
We further verify that our models achieve consistent gains over almost all domains on SGD, as shown in Table~\ref{tab:sgd-domain}. 

Our observations are as follows: 
1) With notably less trainable parameters, SHEGO achieves a prominent and consistent margin over SGD-baseline and Seq2Seq-DU. This indicates that SHEGO can obtain generalized models for DST with higher parameter efficiency. 
2) SHEGO outperforms other parameter efficient approaches, \textit{i.e.} AdapterCL and Prompt-Tuning, suggesting that inducing schema into models effectively improves the performance of learning trainable prompt parameters. 
3) As shown in Table~\ref{tab:sgd-domain}, SHEGO outperforms sixteen out of eighteen domains compared to all baselines under the full-data setting, except for the \textit{``RideSharing''} domain (-4.7 from Prompt-Tuning) and  \textit{``Payment''} domain(-15.4 from AdapterCL). We consider that both have relatively small training size and their slots are different from those of other domains, thus comprising the benefit of scheme graph inducement and multi-domain joint training.
4) Our method has consistent gains by a large margin, especially in domains with a large number of slot types, \textit{e.g.} \textit{``Movies''}, \textit{``Hotels''}, \textit{``Flights''}, suggesting that inducing the schema graph to improve the performance for domains with sufficient slot relations. 

  \begin{figure}[tb]
  \centering
  \includegraphics[width=0.9\linewidth]{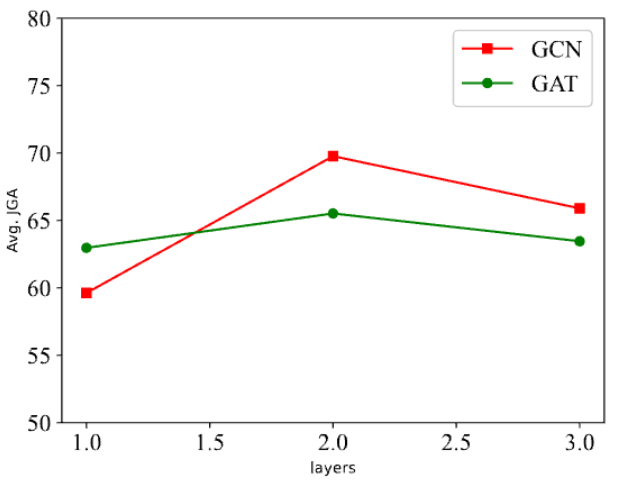}
  \caption{Analysis of the number of pooling layers for GCN and GAT. The hidden size for GNNs is set to 256.}
  \label{fig:gnn}
  \centering
  \includegraphics[width=0.9\linewidth]{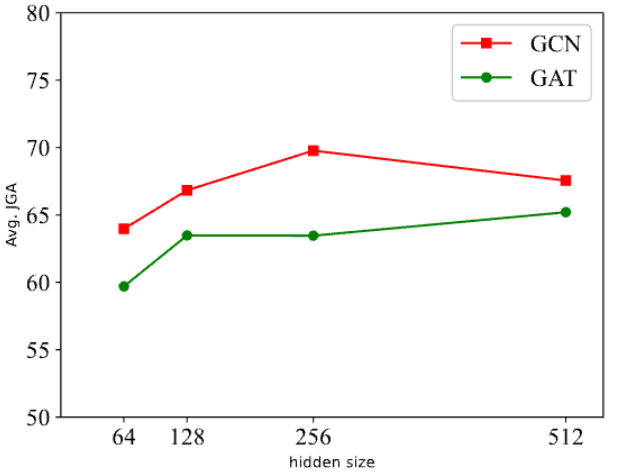}
  \caption{Analysis of hidden size for GCN and GAT. The number of layers is set to two.}
  \label{fig:hidden}
  \label{fig:architect}
\end{figure}

\begin{table*}
  \centering
  \begin{tabular}{c|l|ccc|c}
    \toprule
    \multirow{3}{*}{No.} & \multirow{3}{*}{Model}   & \multicolumn{3}{c}{Applied Strategy}  & \multirow{3}{*}{\textit{Avg. JGA}}\\
    && \multirow{2}{*}{GNN} & Graph & Active & \\
    &&       &Prompts     &Slots & \\
    \midrule
    1&  \textit{w/o Active\&GP}  & \ding{52}   &             &           & 63.5\\
    2&  \textit{w/o GP }& \ding{52}   &             & \ding{52} & 64.6\\
    3&  \textit{w/o Active} & \ding{52}     &  \ding{52}   &            & 67.5\\
    4&  \textit{w/o SlotConnect} &              &  \ding{52}   & \ding{52} & 66.7\\
    5&   Our Model & \ding{52} &\ding{52} & \ding{52} & \textbf{69.8}\\

    \bottomrule
  \end{tabular}
 \caption{Ablation results on the first 15 services of SGD listed in Table~\ref{tab:sgd-services}. All numbers are reported in average joint goal accuracy(JGA) (\%) over 15 tasks. \textit{w/o Active} \textit{w/o Active} means the graph module encodes the same schema information in training. \textit{w/o SlotConnect} indicates the model is trained without graph modules while only adding the same amount of prompt tokens as graph modules.}
 \label{tab:ablation}
\end{table*}

\subsection{Results on MultiWOZ}
\label{sec:mwoz}
Table~\ref{tab:mwoz-joint} shows joint goal accuracy on the MultiWOZ 2.1 test set. 
Our model achieves a competitive JGA of 59.0 on MultiWOZ 2.1 with about 10M tunable parameters. Compared with fine-tuning models, we learn 2 $\sim$ 15 times fewer parameters in SHEGO. This suggests that, instead of fine-tuning, SHEGO still improves the performance of multi-domain DST with induced schema graphs and shared prompt embeddings.
Our model is still somewhat worse than prior DiCoS-DST. This is likely because the DiCoS-DST model utilized ground truth for modeling relations of the slots in the current turn with the last updated turn.
SHEGO also obtains +2.9 on DiSTRICT with the same backbone models, showing that trainable parameters enable greater scalability for our proposed model compared to the reference in-context learning approach.
Prompt Tuning for DST is the closest work to ours, while we use GNN to model structural relations in the schema and promote the flexibility of prompts.

\subsection{Analysis of Schema Graph Architectures}
\label{sec:architect}
In Figure~\ref{fig:architect}, we compare using GAT and GCN in the schema graph and the number of layers for pooling, where the hidden dimension is 256. We evaluate the first 15 services of SGD as listed in Table~\ref{tab:sgd-services} and report the average joint goal accuracy over these services. 
The results show that GCN with two layers outperforms other GNN settings, which is the setup we adopt in our model.
We further study the impact of hidden size for GCN and GAT varying in \{64, 128, 256, 512\}. As illustrated in Figure~\ref{fig:architect}, the hidden size of 256 exceeds others performing the same tasks as above. Based on this, we remark that there is a trade-off between the number of parameters trained from scratch as prompts and the size of the training set.
The results also suggest that even though GNN with more hidden layers can preserve more generalized schema information, generalized schema information might not be fit for prompting from the frozen pre-trained model. 

\subsection{Ablation of Schema Graph Prompt}
\label{sec:abaltion}

To investigate the effectiveness of the embeddings encoded by the schema graph, we conduct an ablation study on our proposed schema graph with variable settings.
Our experiments consist of the following:  
1) w/o Active\&GP turns each graph prompt (GP) into the same token from the model's vocabulary, instead of using graph prompts and masking by the active slots as in our original model. 
2) w/o GP represents substituting each graph prompt with the same token, from the model's vocabulary, and adopts active slot masking for each input example. 
3) w/o Active encodes graph prompts with GNN but is not masked by active slots. 
4) w/o SlotConnect adds shared soft prompts at the same length as graph prompt length and is trained the same with the shared prompts. It is similar to learning parameters of graph prompts in 3) without graph neural network. 

Results are presented in Table~\ref{tab:ablation}. 
First, we observe that formulating input slot nodes as trainable prompts brings about improvement compared to using active slots indices only. Comparing 5) to 2) and 3) to 1), tunable graph prompts enhance Avg. JGA by 4.8 and 4.0. It gives additional trainable parameters to the graph neural network to learn the characteristics of each domain-specific feature.
Second, incorporating active slots in each input example is beneficial to generate masked spans correctly. Comparing 5) to 3) and 2) to 1), masking with active slots help improve JGA by 2.3 and 1.1, respectively. But the gain from indices of active slots is weaker than that from the prompt token formulation.
Third. It is evident to show the effectiveness of our schema graph using a graph neural network comparing 5) with 4). Prompt tokens improve by 3.1 on the accuracy of learning slot connections within the graph neural network. Moreover, prompt tokens cannot benefit from active slots in our empirical experiments, which might be because they make input features too sparse to train an embedder without graph pooling.

\section{Conclusion}

Designing prompts for multi-domain dialogue state tracking is challenging since it involves extra efforts to generate appropriate prompts for dialogue state tracking in different domains.
To address this issue, we propose SHEGO, a graph-based prompt learning method that first incorporates slot relations in the schema and learns domain-aware prompts to apply the domain-specific knowledge as prompts. 
SHEGO encourages the prompts to the pre-trained language model learning shared and domain-specific knowledge at the same time, with other hidden layers in PLM fixed, to exhaust the capabilities of natural language understanding from PLM for dialogue state tracking.
Our experiments on SGD and MultiWOZ 2.1 demonstrate the effectiveness of the schema graph that can capture slot relations with pooling and enables domain adaptation with limited trainable parameters.
Furthermore, SHEGO advances recent parameter-efficient approaches with prompts that involve manual designs or random initialization for domain adaption and contributes to easier and more efficient multi-domain dialogue state tracking.

\section{Limitations}
Our study is subject to four possible limitations. 
First, the input length of dialogue history and the number of total slots are constrained in practice, which means as the length increases, more computational resources are required for maintaining the performance.
Moreover, the information in the dialogue history might lose because of the truncation of previous utterances and is limited by the memory of the device.
Second, training examples per domain are unbalanced in our selected training dataset, which may bring about a bias to domains with high resources, meanwhile affecting the performance in the remaining domains.
Third, we do not include existing work like \citet{gupta2022show} and \citet{yu2020score} in baselines because they apply data augmentation or use extra resources of data that affect evaluation results.
Fourth, we do not present per-domain performance on the MultiWOZ 2.1, because all our baselines either only show the joint goal accuracy on all domains or study under zero-shot/few-shot settings. However, we mainly focus on full-data setups in the paper.
\bibliography{anthology,custom}
\bibliographystyle{acl_natbib}

\appendix

\section{Appendix}
\label{sec:appendix}

\begin{table*}[htb]
  \centering
  \begin{adjustbox}{width=\linewidth}
  \begin{tabular}{lcccccccc}
    \toprule
    Dataset & Domains & \textit{Train} & \textit{Dev} & \textit{Test} & \textit{Train samples} & \textit{Dev samples} & \textit{Test samples} & Avg. Turns \\
    \midrule
    SGD & 18 & 5,278 & 761 & 1,531 & 38,745 & 5,589 & 11,349 & 14.7 \\
    MultiWOZ 2.1 & 5 & 8,324 & 999 & 1,000 &12,426 & 987 & 1,091 & 13.5\\

    \bottomrule
  \end{tabular}
  \end{adjustbox}
   \caption{Statistics of SGD and MWOZ 2.1.}
 \label{tab:stat}
\end{table*}

\begin{table*}
  \centering
  \begin{adjustbox}{width=\linewidth}
  \begin{tabular}{llllllll}
    \toprule
    \multicolumn{2}{l}{\textbf{SGD Services (Training samples)}}  & & & & &\\\midrule
    homes\_1 (1829), &hotels\_1 (868), &rentalcars\_3 (332), &hotels\_3 (737), &media\_2 (215), &hotels\_4 (559), &music\_1 (468), &restaurants\_1 (2098), \\
    rentalcars\_2 (631), &homes\_2 (424), &trains\_1 (415), &flights\_3 (420), &services\_3 (959), &flights\_1 (4680), &services\_4 (680), &flights\_2(822), \\
    movies\_2(118), &music\_3 (112), &media\_3 (327), &ridesharing\_1 (412), &movies\_1 (1873), &rentalcars\_1 (840), &ridesharing\_2 (380), &hotels\_2 (1569), \\
    restaurants\_2 (807), &buses\_1 (1054), &services\_1 (1241), &buses\_3 (405), &alarm\_1 (367), &events\_2 (3537), &music\_2 (857), &movies\_3 (231), \\
    flights\_4 (290), &events\_1 (1424), &weather\_1 (259), &media\_1 (1207), &payment\_1 (233), &services\_2 (917), &hotels\_1 (1593), &hotels\_1 (112), \\
    calendar\_1 (773), &banks\_1 (1138), &events\_3 (220), &banks\_2 (312), & & & & \\
    \bottomrule
  \end{tabular}
     \end{adjustbox}
 \caption{Service names and corresponding domains in Schema-Guided-Dialogue.}
 \label{tab:sgd-services}
\end{table*}

\end{document}